\documentclass[letterpaper, 10 pt, conference]{ieeeconf}  
\usepackage{graphicx}
\usepackage{cite}
\usepackage{url}
\usepackage{booktabs}
\usepackage{multirow}
\usepackage{amssymb}
\usepackage[hidelinks]{hyperref}

\IEEEoverridecommandlockouts                              

\title{\LARGE \bf
Affordance-Aware Interactive Decision-Making and Execution for Ambiguous Instructions
}

\author{Hengxuan Xu$^{1,2,*}$, Fengbo Lan$^{1,2,*}$, Zhixin Zhao$^{1}$, Shengjie Wang$^{3}$, Mengqiao Liu$^{1}$,\\ Jieqian Sun$^{1}$, Yu Cheng$^{4,\dagger}$, Tao Zhang$^{1,\dagger}$,  \textit {Fellow, IEEE} 
\thanks{$^{*}$Equal contributions. $^{\dagger}$Corresponding authors.}%
\thanks{This work was supported in part by the National Science and Technology Major Project under Grant 2022ZD0114903.}%
\thanks{$^{1}$Hengxuan Xu, Fengbo Lan, Zhixin Zhao, Mengqiao Liu, Jieqian Sun, and Tao Zhang are with the Department of Automation, Tsinghua University, Beijing, China.
        {\tt\small \{xu-hx24, lanfb22, zhaozhix22, mq-liu24, sunjq21\}@mails.tsinghua.edu.cn}, {\tt\small taozhang@tsinghua.edu.cn}}%
\thanks{$^{2}$Hengxuan Xu and Fengbo Lan are also with the Shanghai Artificial Intelligence Laboratory, Shanghai, China.}%
\thanks{$^{3}$Shengjie Wang is with the Institute for Interdisciplinary Information Sciences, Tsinghua University, Beijing, China.
        {\tt\small wangsj23@mails.tsinghua.edu.cn}}%
\thanks{$^{4}$Yu Cheng is with the Department of Computer Science and Engineering, The Chinese University of Hong Kong, Hong Kong, China.
        {\tt\small chengyu05@gmail.com}}%
}

\makeatletter
\renewcommand{\@biblabel}[1]{[#1]\hfill}
\makeatother

\begin{document}

\maketitle

\thispagestyle{empty}
\pagestyle{empty}

\begin{abstract}

Enabling robots to explore and act in unfamiliar environments under ambiguous human instructions by interactively identifying task-relevant objects (e.g., cups or beverages for “I’m thirsty”) remains challenging for existing vision-language model (VLM)-based methods, due to inefficient reasoning and limited environmental interaction. These factors hinder real-time task planning and execution. To address this, we propose Affordance-Aware Interactive Decision-Making and Execution for Ambiguous Instructions (AIDE), a human-inspired dual-stream framework that integrates interactive exploration with vision-language reasoning: mirroring the complementary deliberative and intuitive processes of human cognition, Multi-Stage Inference (MSI) serves as the decision-making stream and Accelerated Decision-Making (ADM) as the execution stream, enabling zero-shot affordance analysis and interpretation of ambiguous instructions. Experiments in simulation and real-world settings show that AIDE achieves over 80\% task planning success rate and more than 95\% closed-loop continuous execution accuracy at 10 Hz, surpassing existing VLM-based methods in diverse open scenarios. Video demonstrations of learned behaviors and the code can be found on the \href{https://sites.google.com/view/aide-thu/%E9%A6%96%E9%A1%B5}{supplementary website}.

\end{abstract}

\section{INTRODUCTION}

\begin{figure}[t]
  \centering
  \includegraphics[width=\columnwidth]{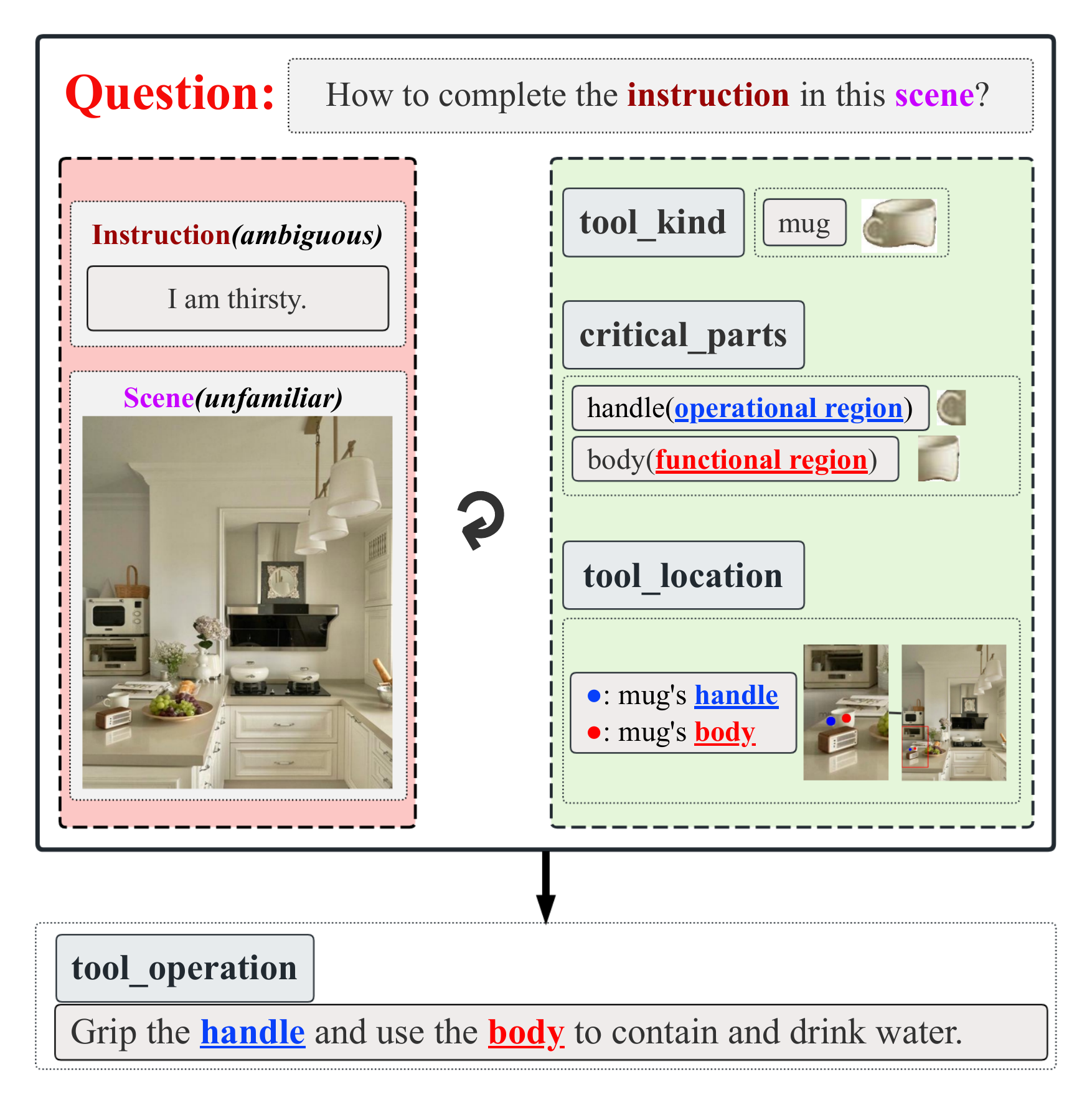}
  \vspace{-14pt}
  \caption{Method objective. Given ambiguous instruction and unfamiliar scene as the input task, the objective of the AIDE is to find the required tool and locate its operational and functional regions within the scene, enabling accurate tool manipulation and task completion.}
  \label{F: Method objective}
\end{figure}

Ambiguous task understanding and object affordance reasoning pose significant challenges for robotic systems \cite{2}, particularly in complex, dynamic environments. Unlike robots, humans naturally interpret ambiguous instructions by evaluating environmental context and adapting actions accordingly, for example, responding to “I am thirsty” (Fig. \ref{F: Method objective}) by fetching a cup, a bottle of water, or a bowl when a bottle is unavailable. Similarly, successful task execution in unfamiliar scenes requires iterative interaction and exploration of both the environment and humans, alongside understanding the affordances and functions of available tools to support effective decision-making. Endowing robots with such human-like strategies is therefore a naturally biomimetic route toward embodied intelligence in open environments.

Recent advances in large language models (LLMs) and vision-language models (VLMs) have demonstrated strong conceptual understanding, common-sense reasoning, and semantic knowledge representation \cite{b9,b10,b32,b53}, encoding extensive world knowledge \cite{65,42,29}. Trained on large-scale Internet data, VLMs can recognize objects and infer their basic functions, for instance, identifying a cup as a drinking vessel. However, their affordance reasoning remains limited: given “I want to crack walnuts,” humans can flexibly substitute a hammer with a heavy book or similarly rigid object when the hammer is unavailable, whereas current VLMs cannot. Furthermore, deploying VLMs for ambiguous task understanding and planning in open-ended environments is prone to hallucinations and uncertainty due to limited environmental interaction, while their high inference cost and latency further constrain applicability to real-world task planning and execution.

\begin{figure*}[t]
  \centering
  \includegraphics[width=\textwidth]{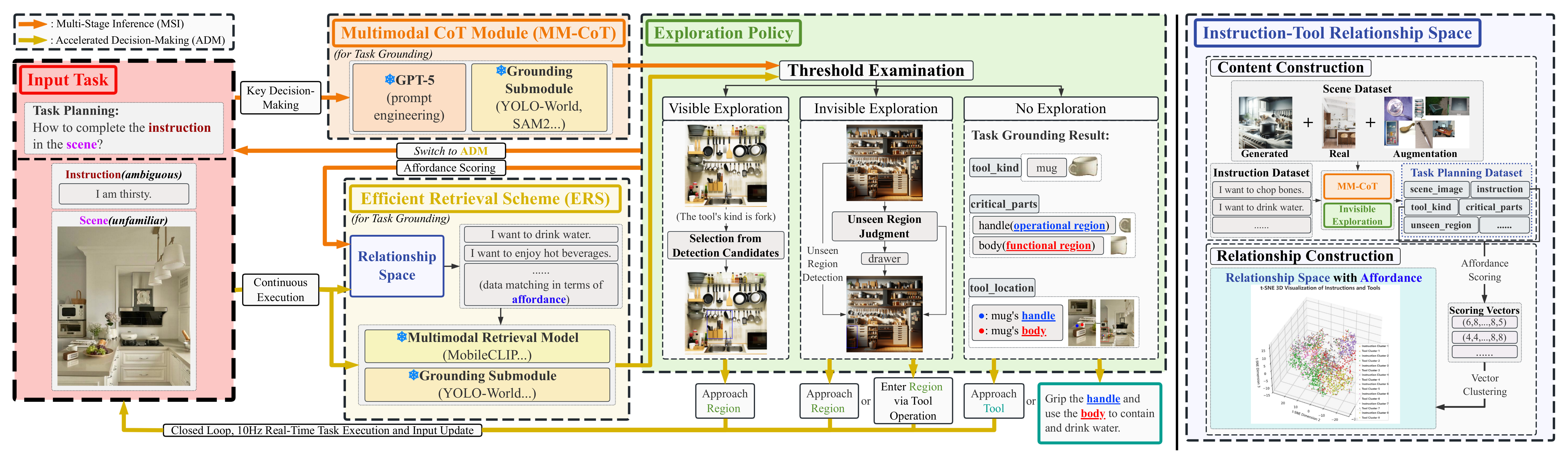}
  \vspace{-10pt}
  \caption{Framework of AIDE. Given the input task, the Multi-Stage Inference (MSI) stream uses Multimodal CoT Module (MM-CoT) and Exploration Policy to generate keyframe-based task planning result, which is projected into the Instruction-Tool Relationship Space via affordance scoring to enhance cross-modal understanding of the input instruction and robustness to GPT-5 hallucinations. Building on this projection, the Accelerated Decision-Making (ADM) stream employs Efficient Retrieval Scheme (ERS) and Exploration Policy for real-time, closed-loop task execution over continuous frames.}
  \label{F: Design of the AIDE method}
\end{figure*}

To address these challenges, we propose AIDE, a novel vision-language task planning framework whose cross-coupled dual-stream mechanism for reasoning and decision-making is inspired by the human coupling of deliberate reasoning with fast, experience-driven decisions (Fig. \ref{F: Design of the AIDE method}). Unlike conventional VLM-based methods, AIDE adopts an efficient closed-loop planning approach through active exploration and interaction with both the environment and human, enabling real-time acquisition, updating, and integration of missing information to significantly enhance task understanding and planning under ambiguous instructions.

AIDE Components and Performance
1) Task Planning Framework.
AIDE employs a human-inspired reasoning-decision dual-stream architecture comprising Multi-Stage Inference (MSI) stream for decision-making and Accelerated Decision-Making (ADM) stream for execution, with four parts: • Multimodal Chain-of-Thought (CoT) Module for zero-shot understanding and planning of ambiguous tasks; • Instruction-Tool Relationship Space for affordance-guided task-tool association; • Efficient Retrieval Scheme for faster and more consistent decision-making; • Interaction Exploration Policy for active exploration and generalization to unseen environments and tasks.
2) Experimental Performance.
AIDE achieves a task-level decision frequency of 10 Hz and over 80\% accuracy across 400 test samples, surpassing state-of-the-art task planning methods and demonstrating its effectiveness as a general-purpose robotic system for ambiguous instruction handling.

\section{RELATED WORK}

\textbf{Visual Language Models and Large Language Models.}
The success of LLMs \cite{b9,29,62} has driven interests in VLMs \cite{23,61,88}, which align the pretrained visual encoder features with LLMs input spaces via cross-modal connectors for single-step end-to-end planning. Research on Chain of Thought (CoT) has further led to the integration of step-wise reasoning and execution capabilities within VLMs and LLMs. Our study focuses on leveraging CoT \cite{wei2022chain} to guide robots in task planning within scenarios, thereby addressing tasks that involve uncertainty.
However, most VLM planning pipelines still lack explicit affordance checks for underspecified human requests.

\textbf{Pre-Trained Foundation Models For Robotics.}
Large pretrained foundation models for robotics span three categories: Pretrained Visual Models, Pretrained Language Models, and Pretrained Visual Language Models. Prior work has explored pretrained models like VLMs for robotics \cite{19,24,39,70}, with RT-2 \cite{8} exemplifying the integration of VLMs in low-level robot control. Methods such as PaLM-E \cite{18} and MOKA \cite{a8} address high-level planning but rely on large-scale visual-language pretraining \cite{11,54}, while VILA \cite{a4} and CoPa \cite{49} target well-specified, single-scenario tasks. In contrast, our work addresses high-level planning related to ambiguous task understanding, planning, and interactive exploration, especially in multi-object, multi-scale open environments.

\textbf{Task and Motion Planning}
\cite{26} integrates low-level motion planning \cite{46} with high-level task planning \cite{22,69}. While classical approaches rely on symbolic \cite{60} or optimization-based methods \cite{77,78}, recent advances in machine learning \cite{41,87}, LLMs \cite{12,16}, and Visual Language Action (VLA) methods \cite{a6,a7} have shifted the field toward end-to-end task execution. Our approach extends VLMs by incorporating object physical properties into a sparse instruction-scene corpus, which encodes rich information such as instructions, scene context, tool attributes, and tool images, to enable efficient task planning with an execution frequency sufficient for closed-loop control.

\section{METHOD}
\label{Method}

We present the AIDE framework (Fig. \ref{F: Design of the AIDE method}), which alternates between two complementary streams, Multi-Stage Inference (MSI) for key decision-making and Accelerated Decision-Making (ADM) for continuous execution, to support accurate planning and real-time closed-loop execution for the input task comprising ambiguous instruction and unfamiliar scene, progressing towards the objective in Fig. \ref{F: Method objective}. The two streams are realized through four parts: the Multimodal CoT Module (MM-CoT) for task grounding, the affordance-aware Instruction-Tool Relationship Space, the Efficient Retrieval Scheme (ERS), and the Interactive Exploration Policy. We next detail each stream and part in turn.

\subsection{Framework}
\label{Streams}
Given the input task, AIDE dynamically alternates between two streams during task execution. This organization mirrors the dual-process structure of human cognition, pairing slow, deliberate reasoning with fast, experience-driven responses. The MSI stream is invoked once for key decision-making when the task is novel or ADM fails to identify any valid tool-related object in the updated scene (specifically, when the sum of the detected tool's detection confidence and its maximum similarity to the retrieved tool images falls below 0.5), producing the task planning result stored for subsequent ADM retrieval. Otherwise, AIDE relies on the ADM stream for continuous execution.

This switching rule is deliberately conservative: MSI establishes the initial semantic anchor when the instruction-scene association is uncertain, whereas ADM is used only when the current scene contains visually and semantically consistent evidence for a grounded tool. Thus, AIDE avoids repeatedly querying the slow reasoning stream for every frame, while still falling back to deliberative reasoning when the scene changes, the target disappears, or retrieved evidence becomes inconsistent with the instruction.

\subsubsection{Multi-Stage Inference Stream}
\label{Multi-Stage Inference}
For the MSI stream, the input task is processed by MM-CoT and the Exploration Policy to produce the task planning result, which is then projected into the Instruction-Tool Relationship Space for storage and indexing. This projection is implemented via DFS-based instruction retrieval in the ERS, which leverages the affordance vector of the input instruction. This process locates the MSI-generated result within the relationship space and retrieves candidate task planning results associated with the input task, supporting cross-modal understanding of the input instruction and robustness to GPT-5 hallucinations for subsequent ADM retrieval and matching.

\subsubsection{Accelerated Decision-Making Stream}
\label{Accelerated Decision-Making} 
The ADM stream enables closed-loop, real-time task execution by continuously performing task planning and updating the scene through robot interaction. Given the input task, ADM performs task planning through the ERS and Exploration Policy, with the former process relying on retrieval from a pool comprising retrieved candidate and MSI-generated task planning results. Operating in a closed-loop manner with real-time input updates at 10 Hz, ADM determines the robot's motion strategy based on exploration outcomes from task planning: (1) visible exploration: the robot approaches the exploration region. (2) invisible exploration: the robot approaches the region when distant or temporarily reformulates the instruction for planning on key regions to enable entry into the exploration region when nearby. (3) no exploration: the robot approaches the tool region when distant or directly manipulates the operational and functional regions to complete the task. The input scene is continuously updated as the robot executes the corresponding motion strategy.

\subsection{Multimodal CoT Module}
\label{Multimodal CoT Module}
The MM-CoT receives the input task and, via prompt-engineered multimodal CoT, jointly leverages GPT-5’s \cite{62} multimodal reasoning and the grounding submodule’s localization capability to produce the task grounding result. It includes labels, images and locations for the required tool and its critical parts. thereby enabling task planning focused on instruction completion within the scene.

Concretely, upon receiving the input task, GPT-5 first predicts the target tool label and its key attributes. This information is passed to the grounding submodule, where the object detection model YOLO-World \cite{a9} and the image segmentation model SAM2 \cite{a10} process the scene to identify the top-$N$ candidate tools by detection confidence. These candidates are returned to GPT-5 for selection of the most appropriate instance's identifier via label and attribute cues. The identifier is then used by SAM2, in coordination with GPT-5, to segment the corresponding tool, distinguishing its operational and functional regions, thus completing the entire multimodal CoT task grounding process.

\subsection{Instruction-Tool Relationship Space}
\label{Instruction-Tool Relationship Space}

\begin{figure}[t]
  \centering
  \includegraphics[width=\columnwidth]{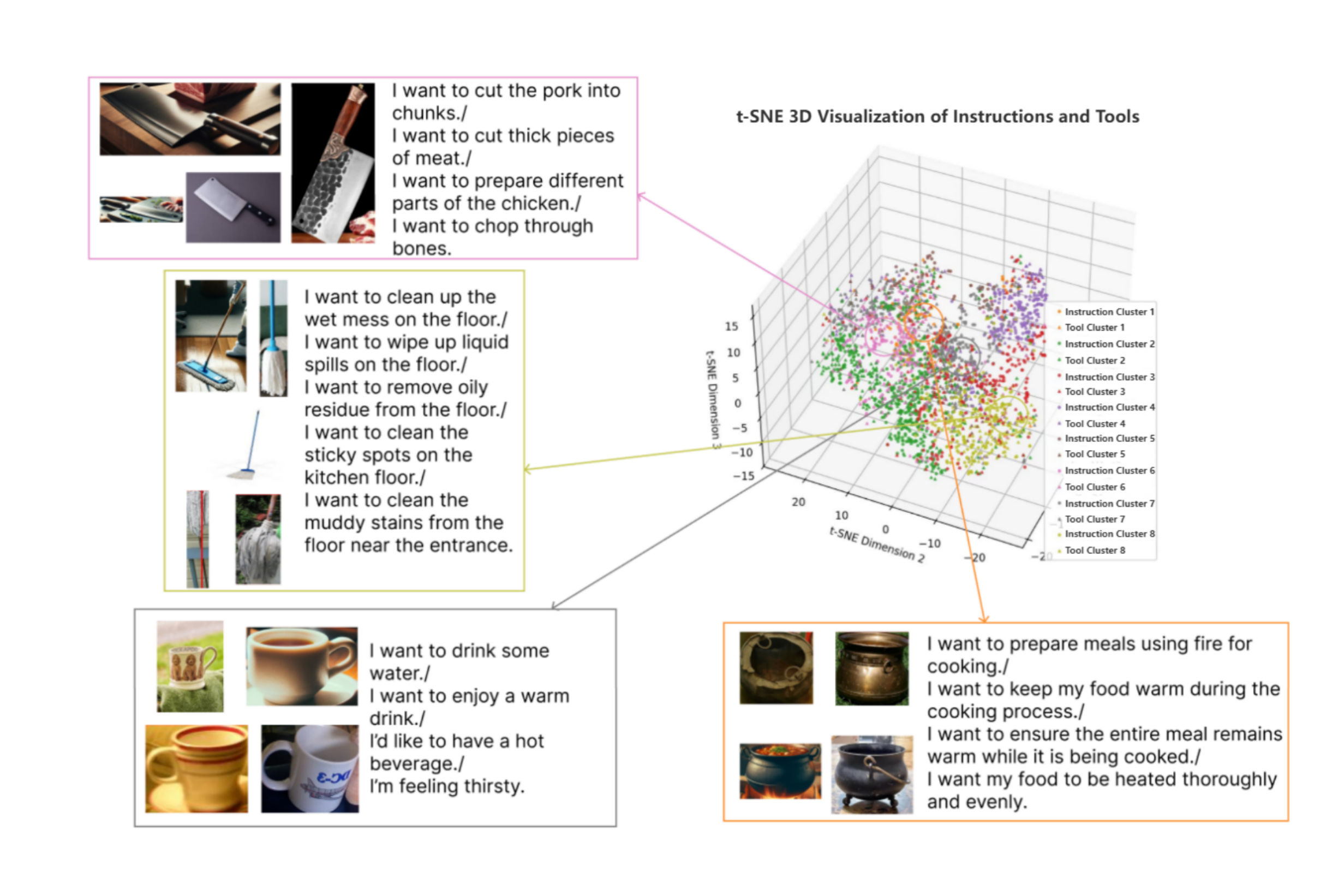}
  \vspace{-14pt}
  \caption{Affordance vector distributions for instructions and tools after t-SNE projection into the three-dimensional space. The image shows that taking cutting/cleaning/drinking/heating-related tasks as representative examples, the affordance vector distributions for instructions and tools largely overlap. }
  \label{F: Clustering distribution of the affordance vectors}
\end{figure}

The Instruction-Tool Relationship Space is built by generating instruction-indexed task planning results as content, and then by establishing many-to-many instruction-tool relationships among them as the primary goal, while also inducing instruction-instruction and tool-tool relationships with functional roles. Accordingly, this space is constructed through two processes: content construction via MM-CoT and the Exploration Policy, and relationship construction from affordance analysis. 

For the former process, diverse instructions and scenes are ensured to yield valid task planning results, which serve as the content. 
We first use GPT-5 to generate $A$ instructions, subsequently rephrased in natural, ambiguous language without explicit tool labels. 
And we construct three scene datasets: (1) DALL-E 3-generated \cite{dalle} scene images (Generated Scene Dataset), (2) real-world scene images containing instruction-relevant tools sourced from the Internet (Real Scene Dataset), and (3) augmentation images combining task-relevant tools with distractors from ImageNet dataset \cite{ImageNet} (Augmentation Scene Dataset). Each instruction is paired with one image from each dataset. For each instruction-scene pair, the task grounding result indexed by the instruction is generated by the MM-CoT, and the label and image of the unseen region are produced by the invisible exploration stage of the Exploration Policy, which will be described later. These results undergo GPT-5-based format validation and manual screening to remove erroneous or hallucinated ones, thereby ensuring each ambiguous instruction indexes three valid task grounding results and unseen regions. Thus, $A$ ambiguous instructions, corresponding three types of scene images, the valid task grounding results, and unseen regions form the Task Planning Dataset, namely the content of the relationship space.

For the latter process, we perform affordance scoring and clustering on each ambiguous instruction paired with tool images in its indexed task planning results. We define affordance for tools as the capacity of their physical properties to satisfy instruction requirements through interaction, and for instructions as the ability tools must have to complete them. Based on this, GPT-5 performs affordance scoring over $X$ dimensions (defined via GPT-5 prompt design) for each instruction and a randomly sampled tool image from its indexed three task planning results in the Task Planning Dataset, yielding $X$-dimensional affordance vectors. These vectors exhibit largely overlapping distributions for instructions and tools, both clustering into $a$ groups via k-means (Fig. \ref{F: Clustering distribution of the affordance vectors}), with most vectors concentrated near cluster centers and a small fraction scattered as noise, potentially due to GPT-5 hallucinations. This indicates affordance vectors scored from instructions and located near cluster centers capture category-level affordances shared by instructions and their required tools, naturally inducing many-to-many instruction-tool, instruction-instruction, and tool-tool relationships across task planning results. We therefore select the $T$ instructions as indexes, whose affordance vectors, together with those of their corresponding tool images, lie within Euclidean distance $D$ of and are closest to their respective cluster centers. We further use k-means to organize them into $a$ clusters with $b$ subclusters each, to facilitate efficient indexing and retrieval. Together, the task planning results indexed by the $T$ instructions and the induced relationships form the whole relationship space.

\subsection{Efficient Retrieval Scheme}
\label{Efficient Retrieval Scheme}

The ERS takes the input task and employs multimodal retrieval models (e.g., MobileCLIP \cite{mobileclip2024}) together with grounding models (e.g., YOLO-World). It performs efficient matching between the input and retrieved candidate task planning results obtained via affordance-based retrieval from the Instruction-Tool Relationship Space. Based on a similarity-based threshold, the scheme outputs either a complete task grounding result or retrieved candidate task planning results.

Specifically, given the input instruction, the scheme performs affordance-based depth-first search (DFS) over clusters in the relationship space until selecting an instruction whose affordance vector lies within Euclidean distance $c$ of the input. It then retrieves instructions from the same sub-cluster whose tool affordance vectors fall within Euclidean distance $d$, each indexing its associated task planning result. Based on these retrieved candidate task planning results indexed from the relationship space, the ERS uses YOLO-World to detect the top-$N$ candidate tools in the input scene ranked by confidence. MobileCLIP is then applied to compute multimodal similarities between candidates and the tool images in the indexed results, and the candidate with the highest similarity $S_{\max}$ (constrained below 1) is retrieved as the required tool. If $S_{\max}$ exceeds $m$, the detection and similarity-matching procedure is repeated within the retrieved tool region to identify operational and functional regions, yielding the complete task grounding result analogous to that of MM-CoT. Otherwise, the ERS outputs the retrieved candidate task planning results for the Exploration Policy.

The thresholded output also provides an interpretable confidence signal for stream switching. A high $S_{\max}$ indicates agreement between visual evidence and retrieved affordance experience, so ADM proceeds with local grounding at frame rate. A low $S_{\max}$ is treated as insufficient evidence rather than immediate failure, and the retrieved candidates are retained as priors for exploration.

\subsection{Exploration Policy}
\label{Exploration Policy}
Given the input task, when $S_{\max}$ exceeds $m$, exploration is skipped and the task grounding result from MM-CoT or ERS is directly output. Otherwise, when the tool is distant or occluded, the Exploration Policy localizes a rectangular region likely containing the tool to support subsequent approach and execution, employing two strategies: visible exploration and invisible exploration. The strategy is selected by a further similarity test: if the maximum similarity between the top $2N$ candidate tools and all retrieved tool images is above 0.75, visible exploration is performed; otherwise, invisible exploration is invoked.

\textbf{Visible Exploration:    }
When the required tool is visible but blurred due to distance, if the required tool is unlikely to be among the top $N$ candidate tools based on the YOLO-World confidence, it is more likely to be found in areas with high confidence ranked from ($N$ + 1) to $N'$, as higher confidence correlates with proximity to ground truth. Accordingly, the interactive visible exploration strategy computes detection weights over $PX$-pixel square regions centered on candidates ranked ($N$ + 1) to 2$N$. The weight for each square region is accumulated based on objects ranked from ($N$ + 1) to $N'$ that intersect or fall within it, with per-object weight calculated as $N'- rank_{conf}$, where $rank_{conf}$ denotes the rank of detection confidence. The minimal bounding rectangle enclosing the highest-weight region and all intersecting objects is selected, thus completing the visible exploration.

\textbf{Invisible Exploration:    }
When the tool is occluded and undetected by YOLO-World, the invisible exploration outputs the unseen region where the tool may exist. When the unseen region list from the retrieved candidate task planning results is available, MobileCLIP selects the unseen region label by matching the input against list labels; otherwise, GPT-5 infers the label from the input. Based on the selected label, YOLO-World retrieves the top $N$ candidate unseen regions by detection confidence. The final exploration region is determined by matching candidates with the unseen regions in the provided list when available, or by GPT-5's selection from the candidate images.

\section{EXPERIMENT}
\label{experiment}
This section presents the experimental design, and analysis of AIDE's results and key characteristics. Through comparisons with state-of-the-art baselines, ablation studies, and error analysis, we validate AIDE's effectiveness and core innovations, namely affordance analysis and exploration policy, in achieving superior task completion accuracy and efficiency.

\subsection{Experimental Setup}
\label{Experimental Setup}

\textbf{a) Baseline:}
We compare AIDE against eight baselines spanning three categories: general MLLMs (Molmo-7B-D-0924 \cite{molmo2024}, Qwen3-VL-30B-A3B-Instruct \cite{qwen3technicalreport}, GPT-5), robot manipulation VLMs/VLAs (RoboPoint \cite{yuan2024robopoint}, Magma-8B \cite{yang2025magmafoundationmodelmultimodal}), robotic task planning methods (SayCan \cite{2}, MOKA, IntroPlan \cite{liangintrospective}).


\textbf{b) Simulation Experiment Setting:}

We empirically set parameters as $A$=1,500, $D$=25, $T$=1,368, $X$=19, $a$=8, $b$=3, $c$=10, $d$=15, $m$=0.85, $N$=5, $N'$=40, and $PX$=250. In simulation, AIDE follows the MSI-ADM process, taking the task as input and producing planning results via ADM over a subsampled Instruction-Tool Relationship Space. This space contains 432 instructions selected from the entire set of $T$=1,368 instructions. The subsample is constructed based on task planning results indexed by 32 cluster-node instructions and 400 randomly sampled instructions, together with their associated relationships. As shown in Table \ref{tab:instruction_naturalness}, instructions in AIDE achieve average naturalness scores above 6 across multiple criteria (higher than IntroPlan and MOKA), confirming their ambiguous yet natural phrasing (Section \ref{Instruction-Tool Relationship Space}). Based on this, we build two test sets, G-Dataset and R-Dataset, by sampling 100 instructions from the remaining 936 instructions in the entire Instruction-Tool Relationship Space and pairing each with two images from the Generated Scene Dataset and the Real Scene Dataset, respectively, yielding 200 instruction-scene pairs per set. The Augmentation Scene Dataset is excluded due to lack of realistic context. The resulting test sets cover diverse tool-visibility conditions, namely clearly identifiable, ambiguous, unrecognizable, or absent, supporting rigorous evaluation of AIDE's closed-loop exploration under realistic indoor settings.

The construction protocol prevents direct overlap between the indexed relationship space used by ADM and the held-out instructions used for testing. Thus, success on the two test sets requires transferring affordance-level relationships to new language expressions and scene layouts, rather than memorizing instruction labels. All baselines are evaluated on the same instruction-scene pairs without task-specific fine-tuning on the test samples.

For evaluation, “TSR”, “OSR”, “FSR” and “WSR” denote tool, operational region, functional region selection, and whole task planning success rate, respectively. “ASR” denotes the task planning success rate in tool-absent scenes, and “FPS” denotes the average execution rate per test sample for each method. “CL”, “AA”, and “RT” represent closed-loop, affordance analysis, and real-time execution functionalities, respectively. All success rates are evaluated via majority vote from three random evaluators.

\begin{table}[t]
\centering
\caption{Instruction Naturalness Analysis. Average naturalness scores (1-10) of instructions from different methods, as evaluated by different LLMs.}
\label{tab:instruction_naturalness}
\vspace{-6pt}
{\scriptsize
\begin{tabular*}{\columnwidth}{@{\extracolsep{\fill}}l r r r@{}}
\toprule
\textbf{Method}
& \textbf{Qwen3-Next-80B}
& \textbf{Llama3-70B}
& \textbf{GPT-5} \\
\midrule
AIDE       & 6.43 & 6.67 & 7.45 \\
IntroPlan  & 3.22 & 4.12 & 5.99 \\
MOKA       & 1.47 & 2.67 & 3.27 \\
\bottomrule
\end{tabular*}
}
\par\vspace{2pt}
{\footnotesize Higher score indicates higher naturalness.\par}
\end{table}

\textbf{c) Real-World Experiment Setting:}

AIDE is deployed on a 6-DOF wheeled robotic arm (a FAIRINO FR-3 arm with an Agibot OmniPicker gripper and a RealSense D435 camera, controlled by a GPU-free industrial computer) and tested in a cluttered indoor environment on six ambiguous instructions labeled 1-6: “I am thirsty,” “I want to clean the dust,” “I want to crack walnuts,” “I want something cold to drink,” “I want to close up delivery boxes tightly,” and “I want to support my waist while sitting.” The robot is evaluated on its ability to navigate to the target tool region in real-time and grasp it appropriately. Here, “Valid Frames” means the number of frames captured by the robot-mounted camera from task initiation to the moment before the successful tool grasp, and “ESR” denotes continuous task execution success rate calculated on valid frames.

During real-world execution, the camera stream is processed while the base and arm move, so the target may change scale, become partially occluded, or leave the field of view for short intervals. We therefore evaluate continuous task execution over valid frames rather than only reporting the final grasp outcome.

\subsection{Simulation Result and Discussion}

\textbf{Comparison Experiments:} 
We evaluate AIDE against eight baselines on the G-Dataset and R-Dataset under zero-shot conditions. As shown in Table \ref{Quantitative Evaluation on Generated Scene Images}, AIDE achieves task planning success rates of 83\% and 88\% on the two datasets, exceeding all baselines by over 30\%. AIDE's tool selection success rates reach 94.5\% and 95.0\%, respectively, outperforming all baselines, which validates the hallucination robustness of MM-CoT in the MSI stream, and the efficiency and well-chosen hyperparameters of the affordance-analysis-based ERS in the ADM stream. Ablations on ERS's retrieval hyperparameters further corroborate these findings: with the DFS termination threshold $c$=10, affordance-based retrieval attains 96.0\% accuracy within 0.0046 s per query, beating exhaustive search (83.0\%) and text-similarity retrieval (at most 87.0\%, over 50× slower). Together, the MSI and ADM designs ensure efficient and reliable tool selection. AIDE also ranks first in operational and functional region selection success rates, collectively validating its effectiveness in task planning under ambiguous instructions in unfamiliar scenes. We further evaluate exploration capability on instruction-scene pairs where target tools are not detectable. The results show that all exploration-capable baselines achieve task planning success rates below 15\%, whereas AIDE exceeds 60\% on both test sets (64.0\% on G-Dataset, 65.0\% on R-Dataset), outperforming other baselines by over 45\%. These results validate the effectiveness of AIDE's interactive exploration policy. For the FPS metric, only AIDE achieves nearly 10 Hz; all baselines except Magma operate below 0.2 Hz. Although Magma attains close to 1 Hz owing to its lightweight design, it yields the lowest task planning success rates on both test sets designed for complex embodied planning. These results demonstrate that AIDE simultaneously achieves high accuracy and efficiency in task planning.

\begin{table*}[t]
\centering
\caption{Zero-shot performance comparison on G-Dataset and R-Dataset.}
\label{Quantitative Evaluation on Generated Scene Images}
\vspace{-6pt}
{\scriptsize
\setlength{\tabcolsep}{4pt}

\resizebox{\textwidth}{!} {
\begin{tabular}{l rrr rrrrr r rrrrr r}
\toprule
\multirow{2}{*}{\textbf{Method}} & \multicolumn{3}{c}{\textbf{Functionality}} 
& \multicolumn{6}{c}{\textbf{G-Dataset}} 
& \multicolumn{6}{c}{\textbf{R-Dataset}} \\ \cmidrule(lr){2-4} 
\cmidrule(lr){5-10} \cmidrule(lr){11-16} & \textbf{CL} & \textbf{AA} & \textbf{RT}
& \textbf{TSR} & \textbf{OSR} & \textbf{FSR} & \textbf{WSR} & \textbf{ASR} & \textbf{FPS}
& \textbf{TSR} & \textbf{OSR} & \textbf{FSR} & \textbf{WSR} & \textbf{ASR} & \textbf{FPS} \\
\midrule

Molmo & & & & 10.5 & 6.5 & 1.0 & 0.5 & 0.0 & 0.05 & 15.5 & 9.5 & 4.0 & 1.5 & 0.0 & 0.05\\

Qwen3-VL & & & & 42.0 & 23.5 & 13.5 & 12.0 & 8.0 & 0.04 & 70.0 & 31.0 & 34.5 & 21.0 & 5.0 & 0.04\\

GPT-5 & & & & 50.5 & 34.5 & 37.5 & 29.5 & \underline{12.0} & 0.03 & 78.0 & 37.0 & 45.5 & 33.5 & \underline{10.0} & 0.04\\ 

\midrule

Magma & & & \checkmark & 1.5 & 0.0 & 0.5 & 0.0 & 0.0 & \underline{1.03} & 1.0 & 0.5 & 0.0 & 0.0 & 0.0 & \underline{0.94}\\

Robopoint & & \checkmark & & 64.0 & 38.0 & 34.0 & 21.5 & 4.0 & 0.14 & 77.5 & 41.5 & 38.5 & 36.5 & 0.0 & 0.19\\

\midrule

SayCan & \checkmark & \checkmark & & 42.5 & - & - & - & - & 0.04 & 59.0 & - & - & - & - & 0.03\\

MOKA & & \checkmark & & 81.5 & \underline{57.5} & \underline{53.5} & \underline{49.0} & - & 0.03 & \underline{88.0} & \underline{68.5} & \underline{59.0} & \underline{54.0} & - & 0.03\\ 

IntroPlan & & & & \underline{82.0} & - & - & - & - & 0.04 & 87.5 & - & - & - & - & 0.04\\

\midrule
AIDE & \checkmark & \checkmark & \checkmark & \textbf{94.5} & \textbf{88.0} & \textbf{89.0} & \textbf{83.0} & \textbf{64.0} & \textbf{9.70} & \textbf{95.0} & \textbf{92.0} & \textbf{91.0} & \textbf{88.0} & \textbf{65.0} & \textbf{9.50} \\
\bottomrule
\end{tabular}
}
}
\par\vspace{2pt}
{\footnotesize \textbf{Bold} indicates the best and \underline{underlined} the second-best performance for each metric. “-” means not supported. All success rates are reported in percentage (\%) except FPS.\par}
\end{table*}

\begin{table*}[t]
\centering
\caption{Ablation study on different components of AIDE.}
\label{Ablation Evaluation on Generated Scene Images}
\vspace{-6pt}
{\scriptsize
\resizebox{\textwidth}{!} {
\begin{tabular}{l rrrrr r rrrrr r}
\toprule
\multirow{2}{*}{\textbf{Method}} 
& \multicolumn{6}{c}{\textbf{G-Dataset}} 
& \multicolumn{6}{c}{\textbf{R-Dataset}} \\
\cmidrule(lr){2-7} \cmidrule(lr){8-13}
& \textbf{TSR} & \textbf{OSR} & \textbf{FSR} & \textbf{WSR} & \textbf{ASR} & \textbf{FPS}
& \textbf{TSR} & \textbf{OSR} & \textbf{FSR} & \textbf{WSR} & \textbf{ASR} & \textbf{FPS} \\
\midrule

MM-CoT & 73.5 & 72.0 & 67.5 & 63.5 & - & 0.03 & 82.0 & 77.5 & 76.0 & 73.0 & - & 0.03\\

ERS & 79.5 & 66.5 & 71.5 & 61.0 & - & \underline{9.50} & 83.0 & 72.0 & 76.0 & 69.0 & - & \textbf{9.80}\\

\midrule

AIDE (w/o ADM) & 90.5 & \textbf{89.0} & \underline{84.5} & \underline{80.5} & \textbf{68.0} & 0.03 & \textbf{96.0} & \underline{91.5} & \underline{90.0} & \underline{87.0} & \textbf{70.0} & 0.03\\

AIDE (w/o MSI) & \underline{92.5} & 79.5 & \underline{84.5} & 74.0 & 52.0 & \textbf{9.70} & 93.0 & 82.0 & 86.0 & 79.0 & 50.0 & \underline{9.50} \\

AIDE (Full) & \textbf{94.5} & \underline{88.0} & \textbf{89.0} & \textbf{83.0} & \underline{64.0} & \textbf{9.70} & \underline{95.0} & \textbf{92.0} & \textbf{91.0} & \textbf{88.0} & \underline{65.0} & \underline{9.50} \\
\bottomrule
\end{tabular}
}
}
\par\vspace{2pt}
{\footnotesize AIDE (w/o ADM) and AIDE (w/o MSI) represent AIDE variants that rely solely on the designed MSI and ADM, respectively, while AIDE (Full) represents the complete framework comprising both MSI and ADM. MM-CoT and ERS denote the corresponding modules operating alone, without the Exploration Policy. Metrics are defined as in Table \ref{Quantitative Evaluation on Generated Scene Images}.\par}
\end{table*}

\textbf{Ablation Experiments:}
Table \ref{Ablation Evaluation on Generated Scene Images} reports ablation results on MSI and ADM in AIDE. Incorporating ADM enables efficient affordance-based retrieval, surpassing MSI-only performance on most metrics, with task planning success rate gains of 2.5\% and 1\% on the G-Dataset and R-Dataset, and a 320× FPS increase to nearly 10 Hz. Incorporating MSI allows AIDE to leverage GPT-5-based scene understanding, yielding 2\%-15\% improvements across success rate metrics. These results demonstrate the distinct yet complementary roles of MSI and ADM in enhancing task planning accuracy and execution efficiency. In addition, rows 1-2 of Table \ref{Ablation Evaluation on Generated Scene Images} ablate other critical components: without the Exploration Policy, MM-CoT and ERS alone cannot handle scenarios requiring exploration of undetectable tools, and all their success rate metrics drop by at least 10\% on both test sets.

\textbf{Functionality Analysis:}
Columns 2-4 of Table \ref{Quantitative Evaluation on Generated Scene Images} present the functional comparison between AIDE and baselines. Among them, SayCan \cite{2} and AIDE support closed-loop planning, while affordance analysis is incorporated into Robopoint, SayCan, MOKA and AIDE. Magma and AIDE enable real-time planning. Specifically, AIDE uniquely achieves real-time (10 Hz) closed-loop planning by combining affordance analysis with the exploration policy, highlighting its advantage over baselines.

\subsection{Real-World Experiments}

\begin{figure}[t]
  \centering
  \includegraphics[width=\columnwidth]{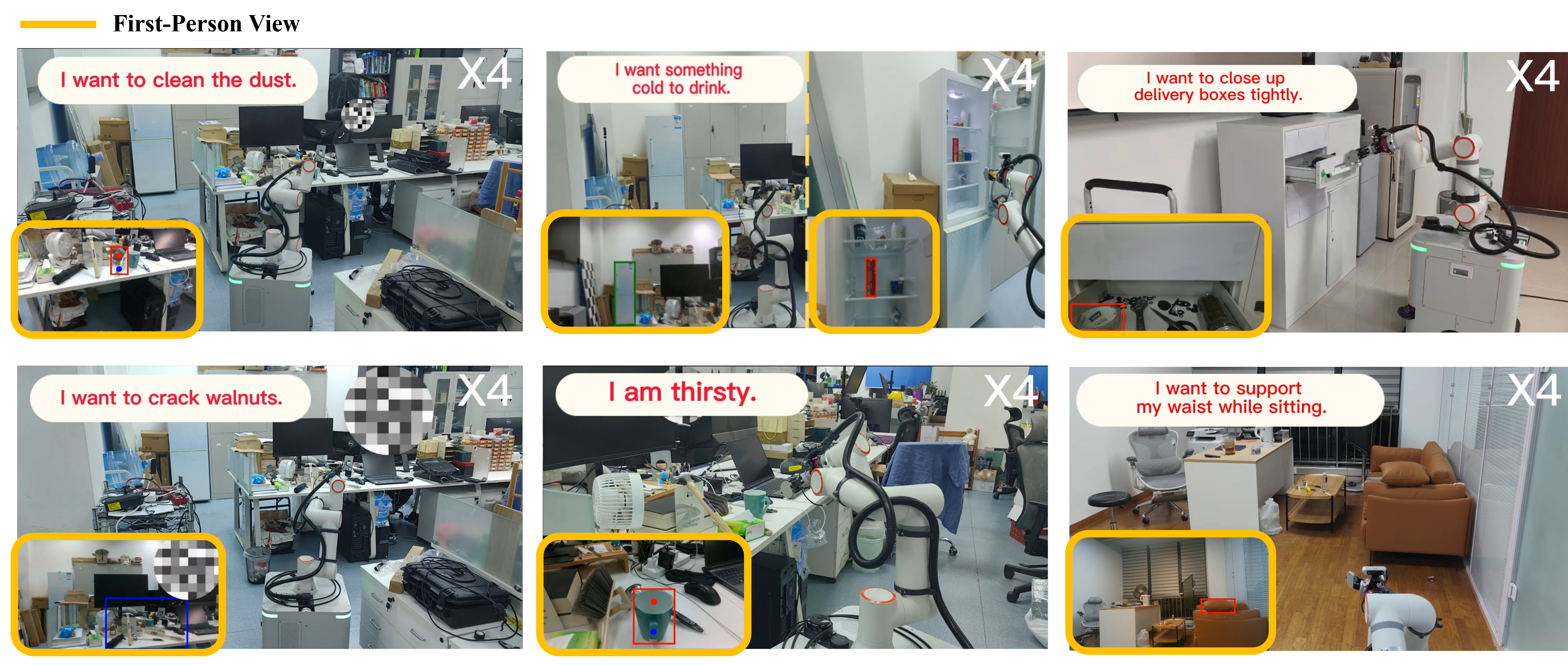}
  \vspace{-14pt}
  \caption{Examples of real-world experiment results. Top left and bottom middle: Identified handle (blue dot) and body (red dot) regions of the target tool. Bottom left: Visible exploration region (blue box). Top middle: Invisible exploration region (green box). Top right and bottom right: tool region (red box).}
  \label{F: Examples of real-world experiment results.}
\end{figure}

As shown in Fig. \ref{F: Examples of real-world experiment results.} and Table \ref{tab:task_planning_accuracy}, the robot successfully performs real-time task planning and execution under Instructions 1-6 via interactive exploration (Fig. \ref{F: Examples of real-world experiment results.}): grasping the cup, brush, hammer, and pillow (Instructions 1-3, 6), and locating the Coke in the fridge and tape in the drawer (Instructions 4-5), with grasping omitted for the latter two instructions given its redundancy with other tasks’ final grasping steps. These results demonstrate the real-time capability and robustness of AIDE's interactive exploration-based closed-loop framework in complex scenarios.

\begin{table}[t]
\centering
\caption{Zero-shot performance on Instructions in real-world experiments.}
\label{tab:task_planning_accuracy}
\vspace{-6pt}
{\scriptsize
\begin{tabular*}{\columnwidth}{@{\extracolsep{\fill}}r l r r@{}}
\toprule
\textbf{No.} & \textbf{Instruction (abbr.)}
& \textbf{Valid Frames}
& \textbf{ESR (\%)} \\
\midrule
1 & Thirsty & 937  & 96.1 \\
2 & Clean dust & 656  & 96.8 \\
3 & Crack walnuts & 1180 & 97.9 \\
4 & Drink cold & 850 & 97.2 \\
5 & Close up boxes & 924 & 97.4 \\
6 & Support waist & 333 & 99.7 \\
\bottomrule
\end{tabular*}
}
\par\vspace{2pt}
{\footnotesize Instruction (abbr.) means abbreviated instructions.\par}
\end{table}

Specifically, in Fig. \ref{F: Examples of real-world experiment results.}, AIDE employs visible exploration to localize distant but observable targets (bottom left), ERS to identify the target tool (top right and bottom right) or its operational and functional regions (top left and bottom middle) upon close approach, and invisible exploration when the target is unlikely to appear in visible regions (top middle), collectively validating the functionalities demonstrated in simulation. Table \ref{tab:task_planning_accuracy} reports AIDE's continuous task execution success rate across six tasks, exceeding 95\% on valid frames in all cases. This result, observed under continuous execution from the robot’s egocentric view, demonstrates the framework's closed-loop robustness in real-world settings and corroborates the task planning success rate in simulation.

\subsection{Error Analysis}

During task planning and execution, AIDE may encounter errors (Table \ref{Quantitative Evaluation on Generated Scene Images}), classified as planning-level or perception-level errors. Planning-level errors stem from incorrect task understanding, triggering ADM to check for valid tool-related objects and switch to MSI if none are found. Failures in MSI are often due to occasional GPT-5 hallucinations, despite the MM-CoT design, and can be mitigated through human interaction for tool re-identification and further exploration. We separately extracted test cases from Table \ref{Ablation Evaluation on Generated Scene Images} where ADM and MSI failed, and evaluated ADM’s error detection rate and MSI’s error recovery rate. Here, EDR (Error Detection Rate) measures the proportion of failed test cases in Table \ref{Ablation Evaluation on Generated Scene Images} for which ADM correctly identifies errors by detecting the absence of valid tool-related objects in the scene. Upon detection, ADM switches to MSI. ERR (Error Recovery Rate) measures the proportion of failed test cases in the same table that MSI successfully recovers through human interaction, including tool reselection and further exploration, without invoking ADM. As shown in Table \ref{Error Analysis}, ADM's detection rate exceeds 50\% and MSI's recovery rate exceeds 40\% across both test sets, validating these mechanisms. Perception-level errors occur when visual degradation from robot motion causes minor area-tool positioning offsets despite correct task understanding. These manifest as sporadic single-frame deviations amid predominantly correct execution frames. As shown in Table \ref{tab:task_planning_accuracy}, such errors account for only 0.3-3.9\% of valid frames without affecting task completion.

\begin{table}[t]
\centering
\caption{Error Detection and Recovery Performance of AIDE.}
\label{Error Analysis}
\vspace{-6pt}
{\scriptsize
\begin{tabular*}{\columnwidth}{@{\extracolsep{\fill}}l rr rr@{}}
\toprule
\multirow{2}{*}{\textbf{Stream}}
& \multicolumn{2}{c}{\textbf{G-Dataset}}
& \multicolumn{2}{c}{\textbf{R-Dataset}} \\
\cmidrule(lr){2-3} \cmidrule(lr){4-5}
& \textbf{EDR} & \textbf{ERR} & \textbf{EDR} & \textbf{ERR} \\
\midrule
ADM & 63.5 & - & 57.1 & - \\
MSI & - & 41.0 & - & 42.3 \\
\bottomrule
\end{tabular*}
}
\par\vspace{2pt}
{\footnotesize Both metrics are reported as percentages (\%).\par}
\end{table}

\section{CONCLUSIONS}




We propose AIDE, a novel human-inspired framework that enables robots to interactively explore and execute tasks under ambiguous instructions in complex indoor environments. Remarkably, AIDE demonstrates superior accuracy and efficiency over state-of-the-art baselines, and effectively supports real-time closed-loop execution in real-world deployment scenarios. By grounding its dual-stream design in human cognitive strategies, AIDE represents a substantial advancement in robot task planning and a step toward biomimetic embodied intelligence empowered by AI, facilitating future research on efficiently handling ambiguous instructions in complex and dynamic environments.

\bibliographystyle{IEEEtran}
\bibliography{IEEEabrv,refs}

\end{document}